\documentclass[12pt]{iopart}
\usepackage{iopams}
\expandafter\let\csname equation*\endcsname\relax
\expandafter\let\csname endequation*\endcsname\relax
\usepackage{amsmath,amsfonts}
\usepackage{algorithm}
\usepackage{algpseudocode}
\usepackage{array}
\usepackage{blindtext}
\usepackage{booktabs}
\usepackage[usenames,dvipsnames]{color}
\usepackage{listings}
\usepackage{rotating}
\usepackage{subcaption}
\usepackage{textcomp}
\usepackage{stfloats}
\usepackage{url}
\usepackage{verbatim}
\usepackage{graphicx}
\usepackage{multirow}  
\usepackage{array}   
\usepackage[binary-units]{siunitx}
\usepackage[backend=bibtex, natbib=true, style=numeric-comp, doi=true, isbn=false, url=false, citestyle=numeric-comp,sorting=none]{biblatex}
\DeclareSIUnit \sop {SOP}
\DeclareSIUnit \timestep {timestep}

\DeclareCaptionSubType*{algorithm}

\DeclareCaptionLabelFormat{alglabel}{Algorithm~#2}

\algnewcommand{\LineComment}[1]{\State \(\triangleright\) #1}

\bibliography{references}

\AtEveryBibitem{%
  \clearfield{note}%
}
\begin{document}

\title[FeNN-DMA]{FeNN-DMA: A RISC-V SoC for SNN Acceleration}
\author{Zainab Aizaz\textsuperscript{\textsection}}
\address{School of Engineering and Informatics, University of Sussex, Brighton, BN1 9QJ, UK}
\ead{z.aizaz@sussex.ac.uk}
\author{James C. Knight$^{*}$\textsuperscript{\textsection}}
\address{School of Engineering and Informatics, University of Sussex, Brighton, BN1 9QJ, UK}
\ead{j.c.knight@sussex.ac.uk}
\author{Thomas Nowotny}
\address{School of Engineering and Informatics, University of Sussex, Brighton, BN1 9QJ, UK}
\ead{t.nowotny@sussex.ac.uk}

\vspace{10pt}
\begin{indented}
\item[]$^*$Corresponding author
\item[] \textsuperscript{\textsection} Equal contribution
\end{indented}


\begin{abstract}
Spiking Neural Networks~(SNNs) are a promising, energy-efficient alternative to standard Artificial Neural Networks~(ANNs) and are particularly well-suited to spatio-temporal tasks such as keyword spotting and video classification.
However, SNNs have a much lower arithmetic intensity than ANNs and are therefore not well-matched to standard accelerators like GPUs and TPUs.
 Field Programmable Gate Arrays~(FPGAs) are designed for such memory-bound workloads, and here we present a novel, fully-programmable RISC-V-based system-on-chip~(FeNN-DMA), tailored to simulating SNNs on modern UltraScale+ FPGAs.
We show that FeNN-DMA has comparable resource usage and energy requirements to state-of-the-art fixed-function SNN accelerators, yet it supports more complex neuron models and network topologies, and can simulate up to 16 thousand neurons and 256 million synapses per core.
Using this functionality, we demonstrate state-of-the-art classification accuracy on the Spiking Heidelberg Digits, Neuromorphic MNIST and Braille tactile
classification tasks.
\end{abstract}

\section{Introduction}
\label{sec:introduction}
Artificial Neural Networks~(ANNs) have demonstrated super-human performance in areas ranging from image classification to language modelling.
However, training current ANNs, and even simply performing inference with them, come at a high energy cost, meaning they face significant limitations in their practical adoption.
The human brain provides a tantalising existence proof that a far more efficient form of neural network is possible, as it runs on only \SI{20}{\watt} and is far more powerful and flexible than any current ANN.
Some of these properties are encapsulated in a biologically-inspired type of ANN known as Spiking Neural Networks~(SNNs), in which individual neurons are stateful, dynamical systems and communicate with each other using spatio-temporally sparse events known as \emph{spikes}.

The main efficiency savings in SNNs come from this event-based communication because, by removing the continuous exchange of activations, the costly matrix multiplication of weights and activations at the heart of ANN computation is replaced by simply adding the weights associated with spiking neurons.
This is particularly effective when spikes are rare events.
However, standard ANN accelerator architectures such as GPUs and TPUs are tailored to the high arithmetic intensity of matrix multiplication, meaning that they are not ideal for SNN acceleration. This has sparked interest in dedicated accelerator architectures for SNNs and, since the 1990s, \emph{neuromorphic engineers} have sought to develop hardware, better suited to accelerating these spiking models.
The first silicon spiking neurons~\citep{mahowald_silicon_1991} were developed using sub-threshold analog circuits and
some neuromorphic systems continue to be built in this way~\citep{richter_dynap-se2_2024}.
However, the majority of modern large-scale systems are implemented as purely digital Application Specific Integrated Circuits (ASICs)~\citep{true_north,Davies2018,spinnaker,gonzalez_spinnaker2_2024}. 
This not only simplifies design but also enables programmable neurons and synapses to be implemented~\citep{Davies2018,spinnaker,gonzalez_spinnaker2_2024}.

Since the early 2000s, there has also been ongoing interest in using Field Programmable Gate Arrays~(FPGAs) to accelerate SNNs (see \citet{mehrabi_fpga-based_2024} for a thorough review).
When compared to ASIC devices manufactured on the same process node, the programmability of FPGAs adds a significant overhead in terms of silicon area, which, in turn, results in higher power consumption and lower clock speeds~\citep{Kuon2006,boutros_you_2018}.
However, FPGAs have practical advantages. 
FPGA design cycles are much shorter, they are available, pre-packaged in a variety of form-factors and FPGA vendors have access to newer process technologies than those typically used to manufacture neuromorphic ASICs, which reduces the overheads.
Furthermore, for memory-bound workloads like SNN simulations, the bottlenecks are likely to be the throughput of the off-chip memory controller and the capacity of the on-chip memory capacity -- both of which are implemented using ASIC-like `hard-blocks' in modern FPGA architectures.

Several large-scale SNN accelerators have been developed using multiple, interconnected FPGAs~\citep{bluehive,deepsouth,neuroaix}, notably DeepSouth~\citep{deepsouth}, which is currently the largest neuromorphic system in the world.
There are also a plethora of FPGA-based SNN accelerators specifically designed for convolutional SNNs~\citep{li_firefly_2024,
chen_sibrain_2024} and on-chip learning using Spike-Timing Dependent Plasticity~(STDP)~\citep{euler_runge,morphbungee}.
Here, we focus on inference with non-convolutional SNNs~\citep{dvs,fpga_nhap,essa,spiker+,li_fully-parallel_2024}.
Because the propagation of spikes tends to be the most time-consuming part of SNN simulation and it is very memory-bound, the throughput of all these systems is essentially constrained by the clock speed~($f_\text{max}$) and how much parallelism is available to accumulate synaptic weights~($N_\text{P}$).
With a fully-pipelined design, this means the theoretical peak throughput of most of these systems is $N_\text{P} \times f_\text{max}$.
Some systems~\citep{li_fully-parallel_2024,spiker+} have dedicated circuits for each neuron but, while this approach enables high throughput, it does not scale to larger models. 
Instead, the majority of systems~\citep{dvs,fpga_nhap,essa} use time-multiplexing to distribute updates of `virtual' neurons and synapses across a smaller number of Processing Elements~(PEs), allowing resource usage to be traded off against throughput. 

Biological neural networks have sparse connectivity~\citep{Perin2011} and several systems~\citep{dvs,essa,li_fully-parallel_2024} implement some form of weight matrix `compression', enabling them to `skip' over the many zeros present in sparse connectivity matrices. This improves the effective throughput and reduces memory bandwidth demands.
As well as accumulating the weights associated with incoming spikes, SNN accelerators also need to update the state of each neuron.
The majority of systems update all neurons every simulation timestep but, \citet{dvs} implemented a fully event-driven neuron update pipeline.
With Leaky Integrate-and-Fire neurons and instantaneous synapses, these event-based updates can be implemented in an elegant and hardware-friendly manner, but, as \citet{Brette2007} discuss, adding non-instantaneous synapses, delays or recurrent connectivity all require significant additional complexity, so that this approach is unlikely to generalise to more complex neuron models.

In order to reduce power consumption and improve throughput, all of the systems discussed above aside from NHAP~\citep{fpga_nhap}, store weights and neuron states in on-chip BlockRAM~(BRAM), meaning that many systems do not support enough neurons or synapses to implement state-of-the-art models (see \citet{fabre_structured_2025} for static and dynamic memory requirements of a range of models).
The NHAP system~\citep{fpga_nhap} does support external memory, but it directly streams data from external memory to its PEs, meaning that memory latency significantly reduces throughput ~(\SI{0.019}{\giga\sop\per\second}).

The existing FPGA systems discussed above include numerous novel architectural features and show impressive performance, but they are almost exclusively tailored to the classification of image-based datasets (primarily MNIST) using networks converted from ANNs. 
As \citet{davies_advancing_2021} showed in their prominent survey of applications benchmarked on Intel's Loihi neuromorphic system~\citep{Davies2018}, this is not an efficient use of SNNs, and the benefits of neuromorphic hardware over ANN accelerators in tasks of this sort are minimal.
Instead, state-of-the-art SNN research focuses on training SNNs directly on more challenging spatio-temporal datasets such as the Spiking Heidelberg Digits~\citep{Cramer2020} or Neuromorphic-MNIST~\citep{orchard_converting_2015}.
State-of-the-art SNNs often feature recurrent connectivity~\citep{baronig_advancing_2024,meszaros_efficient_2025}, synaptic delays~\citep{meszaros_efficient_2025,hammouamri_learning_2023} and significantly more complex neuron models~\citep{baronig_advancing_2024,fabre_structured_2025} than those supported by the systems described above, suggesting that more flexible accelerators are required.
\citet{spiker+} provide an interesting solution to this problem by developing a framework for \emph{generating} task-specific FPGA-based SNN accelerators rather than developing a single accelerator design.
\citet{gomes_de_farias_neurohls_2025} employ a similar approach but generate code in a subset of C suitable for High-Level Synthesis~(HLS) rather than in a Hardware Description Language~(HDL).
However, both approaches require the end-user to synthesize the generated accelerators using FPGA tools 
which places a high entry barrier for users.

Another solution is to build \emph{programmable} FPGA accelerators similar to the large-scale ASIC systems discussed at the beginning of this section.
Because they allow one set of control logic to be shared between multiple parallel ALUs, Single Instruction Multiple Data~(SIMD) or vector architectures are a popular choice for such systems.
\citet{Naylor2013} built a \SI{256}{\bit} wide vector co-processor for a NIOS II CPU and demonstrated that it was a resource-effective way of saturating the external memory bandwidth of an Altera Stratix IV FPGA -- a key goal for any accelerator targeting memory-bound workloads. 
More recently, \citet{chen_gaban_2022} built a vector processor architecture on an AMD UltraScale+ FPGA with High Bandwidth Memory~(HBM). 
\citet{sripad_snavareal-time_2018} took a somewhat different approach and developed an entirely bespoke architecture with a programmable `controller' which implements the limited control flow required by SNNs and offloads the execution of SIMD arithmetic instructions to an array of PEs, each with its own bank of BRAM.

In parallel with the development of specialised FPGA and ASIC-based SNN accelerators, the open-source RISC-V architecture has caused a broader revolution in processor design. 
Not only have organisations such as the OpenHW Foundation made verified open-source cores ranging from microcontrollers~\citep{davide_schiavone_slow_2017} to superscalar application class cores~\citep{tedeschi_cva6s_2025} freely available but the RISC-V Instruction Set Architecture~(ISA) was designed from the ground up to be extendable, making RISC-V an ideal springboard for accelerator designs.
Several standard RISC-V extensions have been developed, including a general-purpose vector extension~\citep{RVV}.
However, standard extensions can struggle in some applications~\citep{SPEED}, due to inefficiencies in handling key computations and because of the size and complexity of the extensions.  
Therefore, numerous specialised RISC-V accelerators have been developed for applications including robotics~\citep{riscv_robotics}, cryptography~\citep{riscv_cryp} and AI~\citep{rsc5_cnn1}.
Perhaps unsurprisingly, several RISC-V-based accelerators have also been developed for SNNs~\citep{rsc5_snn2,rvscnn,spikestream}, although we are not aware of any designed for FPGA deployment.
\citet{rvscnn} pair a four-stage pipelined RISC-V core with a  $4\times4$ array of PEs to accelerate spiking and non-spiking CNNs. \citet{spikestream} also focus on spiking CNNs using a cluster of `Snitch' cores~\citep{zaruba_snitch_2021}, supporting narrow \SI{64}{\bit} floating point SIMD operations alongside `streaming registers' and sparsity extensions to improve the performance on memory-bound workloads.
\citet{yang_back_2023} accelerated both convolutional and recurrent SNNs using a system which pairs a pipelined RISC-V core with 1024 dedicated PEs with their own load-store unit and scratchpad memories.
Finally, \citet{rsc5_snn2} uses multiple lightweight `Zero-riscy' cores~\citep{davide_schiavone_slow_2017} supporting narrow \SI{32}{\bit} SIMD operations, which can be used to implement LIF neurons in just a few instructions.

In our previous paper~\citep{aizaz_fenn_2025}, we presented the first prototype of FeNN -- a RISC-V-based vector processor designed to accelerate SNNs on FPGA.
Here we present FeNN-DMA -- a complete System-on-Chip design based around an extended version of our FeNN core. The main contributions of this work are:

\begin{figure}[b]
    \centering
    \includegraphics{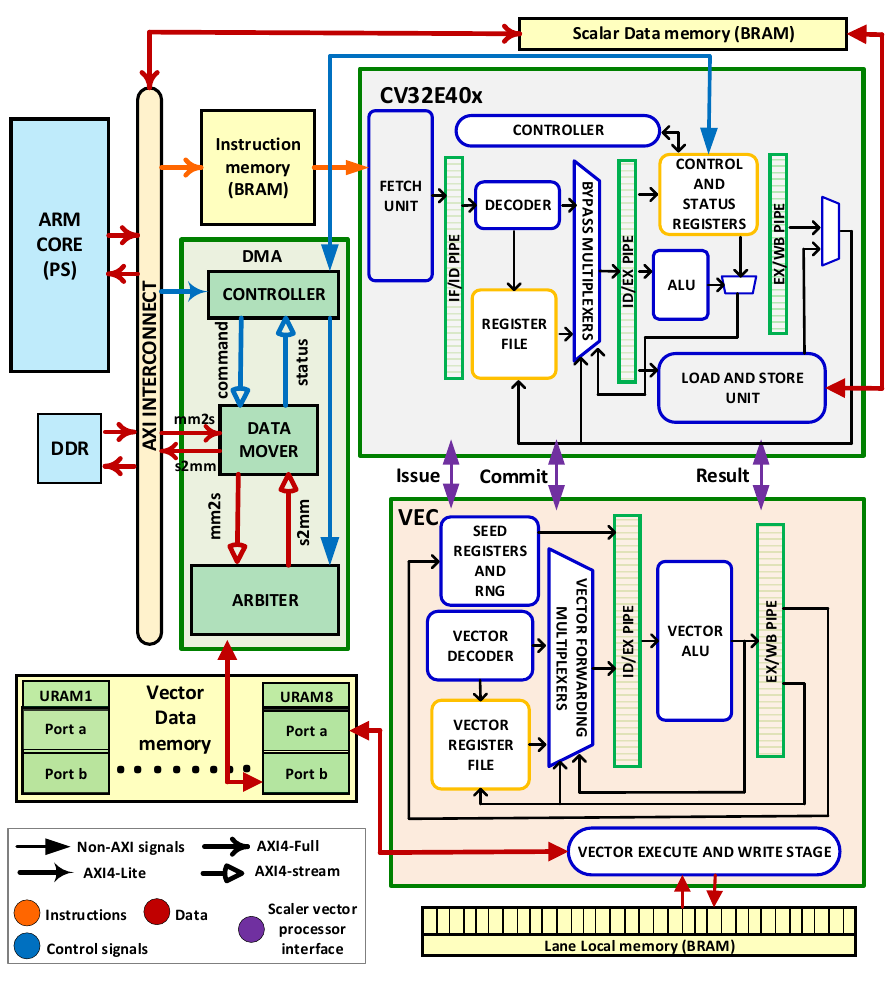}
    \caption{Block diagram of a single-core FeNN System-on-Chip.}
    \label{fig:soc}
\end{figure}
\begin{enumerate}
    \item An updated version of the prototype FeNN core presented in our previous work~\citep{aizaz_fenn_2025} with lane-local memories to support SNNs with synaptic delays and sparse connectivity, as well as additional instructions to optimise key computations.
    \item A bespoke Direct Memory Access~(DMA) controller for efficiently streaming weights and copying simulation data between off-chip and on-chip memory. 
    \item A Python-based programming framework, allowing neuron models to be defined in a C-like language and SNNs to be defined using PyTorch-like syntax or imported from the Neuromorphic Intermediate Representation~\citep{pedersen_neuromorphic_2024}.
    \item Single and dual-core FeNN-DMA system-on-chip (SoC) designs capable of simulating 16 thousand neurons with 256 million synapses on each core, deployed on a Kria KV260 development kit and running state-of-the-art SNN classifiers for the Neuromorphic MNIST~\citep{orchard_converting_2015}, Spiking Heidelberg Digits~\citep{Cramer2020} and Braille tactile classification~\citep{muller-cleve_braille_2022} benchmarks.
\end{enumerate}

\section{Method}
\subsection{Design and implementation of System-on-Chip}
\label{sec:soc_design_implement}
We implemented the FeNN-DMA SoC shown in \figurename~\ref{fig:soc} on a AMD K26 System-on-Module~(SoM) which combines an ARM processor -- referred to as the Processing Systems~(PS) -- with UltraScale+ FPGA fabric -- referred to as the Programmable Logic~(PL) -- and \SI{4}{\giga\byte} of DDR4~(Double Data Rate) memory.
Components in these systems are typically connected together using the AXI (Advanced eXtensible Interface) family of interfaces, and \figurename~\ref{fig:soc} shows how three different types of AXI interfaces are used to connect the components that make up our FeNN-DMA design. 
\emph{AXI4-Full} is a high-performance, memory-mapped interface, supporting burst-mode for high-throughput transfers. 
\emph{AXI4-Lite} is a simpler, memory-mapped interface, typically used for interacting with memory-mapped registers, and \emph{AXI4-stream} is a point-to-point streaming interface.
The following sub-sections describe the system in more detail.

\subsubsection{ARM core}
The ARM core is a quad-core ARM Cortex-A53 processor responsible for running high-level operating systems (like Linux), managing peripherals and controlling application-level tasks. 
In the proposed design, the ARM core runs SNN applications implemented using our PyFeNN library (Section~\ref{sec:methods/pyfenn}) on Ubuntu Linux. 
PyFeNN compiles network descriptions into RISC-V code and data, which gets written to memory-mapped on-chip BRAM or DMA buffers located in the DDR memory. PyFeNN then controls simulations by interacting with the FeNN cores via memory-mapped resources.  

\subsubsection{On-chip memories}
\label{sec:methods/on_chip_memory}
On-chip memories are embedded within the PL and offer low-latency, high-bandwidth storage close to the computational elements.
The AMD UltraScale+ FPGA architecture provides two on-chip memory primitives --  Block RAMs~(BRAMs) which can be used in either \SI{18}{\kilo\bit} or \SI{36}{\kilo\bit} configurations and support data widths of \SIrange{1}{72}{\bit} and Ultra RAMs~(URAMs) which provide a fixed $4096 \times \SI{72}{\bit} = \SI{288}{\kilo\bit}$ configuration. Both types of memory have two ports and a minimum read latency of 1 clock cycle. As shown in \figurename~\ref{fig:soc}, FeNN uses these primitives to implement several different memories, tailored to efficient SNN execution. 
\begin{description}
\item[Instruction and scalar data memories] Because FeNN uses a Harvard architecture, we use separate \SI{32}{\bit} wide memories for instructions and scalar data, implemented using BRAMs and accessible from the PS via AXI BRAM controller IP blocks.
\item[Vector data memory]  Previous FPGA-based accelerators have primarily used BRAM memories as they better suit the typical architectures of multiple Processing Elements operating independently on narrow data types.
However, because FeNN is a wide SIMD processor, we take advantage of the higher-capacity, denser URAM blocks available on UltraScale+ and implement large, on-chip \emph{vector memories} for weights and neuron state using 8 parallel banks of URAM.
\item[Lane Local memory] While the vector memories can handle the access patterns required to update neurons and propagate spikes through dense connectivity without delays, to support delays and sparse connectivity, we require \emph{indexed load} instructions where each lane can access an independent address.
We therefore use one \SI{18}{\kilo\bit} BRAM per-lane to implement a \SI{16}{\bit} wide \emph{lane-local memory} which, as
\citet{Naylor2013} showed, can be used to efficiently implement SNNs with sparse connectivity.
Without indexed load support, SIMD processors have to serialise this type of operation across multiple cycles~\citep{chen_gaban_2022}, reducing performance.
\end{description}

\subsubsection{DDR memory and AXI interconnect}
\label{sec:methods/external_memory}
Although the PL of the K26 SOM has \SI{2}{\mega\byte} of UltraRAM, which is sufficient to store the state of hundreds of thousands of typical spiking neurons, it is not sufficient for storing the synaptic weights of large state-of-the-art networks.
Luckily, the K26 SOM comes with \SI{4}{\giga\byte}
of DDR4-2400 memory for high-speed (peak bandwidth of \SI{18.75}{\giga\byte\per\second}) off-chip data storage. 
The DDR4 memory controller is connected to the PL through four `high-performance' AXI slave interfaces.
These interfaces have a configurable width of up to \SI{128}{\bit} and our experiments suggest that they can deliver data to the PL at a clock speed of over \SI{328}{\mega\hertz} equating to a throughput of around \SI{4.9}{\giga\byte\per\second} per-interface.
This is over $3\times$ the memory throughput achieved by older systems like NHAP~\citep{fpga_nhap}.
In FeNN-DMA, each core's DMA controller is connected to one of these interfaces, using an AXI SmartConnect block to handle crossing from the \SI{328}{\mega\hertz} clock domain to the \SI{175}{\mega\hertz} domain used for FeNN and widening the AXI transactions to match FeNN's \SI{512}{\bit} vector width.
By quadrupling the width and halving the clock speed, the DMA controller receives a vector every two clock cycles, providing a good match for the tightest spike processing loop.

\begin{table}[t]
 \centering
\begin{tabular}{cl}
\textbf{Instruction} & \textbf{Operation}\\
\toprule
VLUI(\textbf{rd}, imm) & \textbf{rd}[i] = imm \\
\midrule
  VADD(\textbf{rd}, \textbf{rs1}, \textbf{rs2})  & \textbf{rd}[i] = sat(\textbf{rs1}[i] + \textbf{rs2}[i])  \\

  VSUB(\textbf{rd}, \textbf{rs1}, \textbf{rs2}) & \textbf{rd}[i] = sat(\textbf{rs1}[i] - \textbf{rs2}[i]) \\

  VAND(\textbf{rd}, \textbf{rs1}, \textbf{rs2}) &\textbf{rd}[i] = \textbf{rs1}[i] \& \textbf{rs2}[i]\\

  VSL(\textbf{rd}, \textbf{rs1}, \textbf{rs2}) &\textbf{rd}[i] = \textbf{rs1}[i] $<<$ \textbf{rs2}[i] \\

 VSR(\textbf{rd}, \textbf{rs1}, \textbf{rs2}) &\textbf{rd}[i] = \textbf{rs1}[i] $>>$ \textbf{rs2}[i] \\

  VMUL(\textbf{rd}, \textbf{rs1}, \textbf{rs2}, shift) &\textbf{rd}[i] =  round(\textbf{rs1}[i] * \textbf{rs2}[i]) $>>$ shift \\
\midrule
  VTEQ(rd, \textbf{rs1}, \textbf{rs2})  & rd[i] = \textbf{rs1}[i] $==$ \textbf{rs2}[i]\\
 
  VTNE(rd, \textbf{rs1}, \textbf{rs2})  &rd[i] = \textbf{rs1}[i] $!=$ \textbf{rs2}[i] \\

  VTLT(rd, \textbf{rs1}, \textbf{rs2}) &rd[i] = \textbf{rs1}[i] $<$ \textbf{rs2}[i] \\

  VTGE(rd, \textbf{rs1}, \textbf{rs2})  &rd[i] = \textbf{rs1}[i] $>=$ \textbf{rs2}[i] \\
\midrule
VSEL(\textbf{rd}, \textbf{rs1}, \textbf{rs2}) & \textbf{rd}[i] = rs1[i] $?$ \textbf{rs2}[i] $:$ \textbf{rd}[i]\\
\midrule
  VSLI(\textbf{rd}, \textbf{rs1}, shift)& \textbf{rd}[i] = \textbf{rs1}[i] $<<$ shift  \\

  VSRI(\textbf{rd}, \textbf{rs1}, shift) & \textbf{rd}[i] = round(\textbf{rs1}[i]) $>>$ shift \\
\midrule
   VRNG(\textbf{rd})  & \textbf{rd}[i] = rng() $>>$ 1  \\
  VANDADD(\textbf{rd}, \textbf{rs1}, rs2, shift) &\textbf{rd}[i] = (\textbf{rs1}[i] \& ((1 $<<$ shift) - 1)) + rs2
 \\
  
\midrule
  VLOAD.V(\textbf{rd}, rs1, offset) & \textbf{rd}[i] = VMEM[((rs1 + offset) / 2) + i]\\

 VLOAD.L(\textbf{rd}, \textbf{rs1}, offset) &  \textbf{rd}[i] = VLOCALi[(\textbf{rs1}[i] + imm) / 2]\\

  VLOAD.R0(rs1, offset) & \textbf{seed\textsubscript{0}}[i] = VMEM[((rs1 + imm) / 2) + i] \\

  VLOAD.R1(rs1, offset) &\textbf{seed\textsubscript{1}}[i] = VMEM[((rs1 + imm) / 2) + i]\\
\midrule
VEXTRACT(rd, \textbf{rs1}, index)  & rd = \textbf{rs1}[index]\\

  VFILL(\textbf{rd}, rs1) &\textbf{rd}[i] = rs1\\
\midrule
VSTORE.V(rs1, \textbf{rs2}, offset) & VMEM[((rs1 + imm) / 2) + i] = \textbf{rs2}[i]\\

VSTORE.L(\textbf{rs1}, \textbf{rs2}, offset) & VLOCALi[(\textbf{rs1}[i] + imm)  / 2] = \textbf{rs2}[i]\\
\bottomrule
\end{tabular}
\caption{Details of the custom vector instructions of FeNN-DMA. `rs1' and `rs2' indicate the source register operands for an instruction and `rd` the destination register operand. Bold indicates registers in the vector register file. `i' represents the index of the vector lane.}
\label{tab:cust_instr}   
\end{table}
\subsubsection{DMA controller}
\label{sec:methods/dma}
Using a DMA controller to copy data from external to internal memory allows the latency and transfer time of external memory while the FeNN core processes previously copied data.
SpiNNaker~\citep{spinnaker} takes a similar approach, using interrupts triggered by the DMA controller to switch context between updating neurons using data in internal memory and processing weights transferred from external memory.
However, because FeNN is a vector processor, its register file is larger than that of a scalar core, so interrupt-based context switching would be prohibitively expensive.
As illustrated in \figurename~\ref{fig:soc}, FeNN's DMA controller is composed of three components, an AMD AXI DataMover IP, and bespoke \emph{CONTROLLER} and \emph{ARBITER} modules. 
The DataMover performs high-throughput data transfers between the AXI memory-mapped and stream protocols. 
In the controller module, we implemented a Finite State Machine~(FSM) to issue commands to the DataMover and to assess the status of the transaction. 
Other parts of the system interact with the DMA controller via registers, which are both memory-mapped via an AXI-lite slave (so they can be accessed from the PS) and exposed to FeNN as custom RISC-V Control \& Status Registers~(CSRs). 
The Arbiter module arbitrates between the MM2S~(Memory-Mapped to Stream) and S2MM~(Stream to Memory-Mapped) DataMover ports and the second port of the URAM-based vector memory. 
The arbiter also handles the generation of URAM addresses and the distribution of stream data across the parallel URAMs that make up the vector memory using another FSM. 

\subsubsection{Scalar Core (CV32E40x)}
\label{sec:methods/cv32e40x}
Tightly coupled co-processors like FeNN allow co-processor instructions to be freely mixed with standard RISC-V, supporting more complex control flow and algorithms that use both processors.
However, the performance of a tightly-coupled co-processor relies on a processor able to issue a co-processor instruction every cycle.
For example, BlueVec~\citep{Naylor2013} used an Altera NIOS II softcore processor, which was not able to do this, requiring the addition of a separate instruction reply mechanism.
To address this issue, we used a CV32E40X~\citep{cv32e40x} core developed by the OpenHW Group. It is a \SI{32}{\bit}, in-order RISC-V processor that supports the baseline RV32I instruction set as well as some standard extensions (we use M and B).
The core is implemented as a four-stage pipeline, enabling it to issue one instruction per cycle in the absence of hazards, and also features an extension interface, making it ideal for hosting our proposed vector co-processor.
However, the CV32E40X was originally designed for ASIC synthesis and, as such, the instruction fetching and load and store units use a standard OBI asynchronous memory interface. For FPGA implementation, we have simplified and optimised these modules to work with synchronous BRAM memories. 

\subsubsection{Vector Core (VEC)}
\label{sec:methods/vec}
The proposed vector processor (VEC) is a 3-stage pipelined (decode, execute and writeback), 32-lane processor with a vector width of \SI{512}{\bit}. The CV32E40x offloads instructions to VEC for decoding through the `issue' interface and signals the execute stage whether an issued instruction should be committed (i.e. its effects made permanent) or killed through the `commit' interface. Finally, the `result' interface is used to transfer scalar results back to the CV32E40x register file. VEC has a $32\times\SI{512}{bit}$ register file  with two read ports and one write port, implemented using distributed memory. Data hazards occurs when one instruction depends on the result of a previous unfinished instruction and, to prevent these causing stalls, VEC implements bypass multiplexers for its vector registers.
These detect whether the operand data for the instruction being decoded is available in the output from the Arithmetic and Logic Unit~(ALU) and, if so, forwards it directly to the decode stage instead of reading it from the register file. 
We have removed compressed instruction support from the CV32e40x core and used an entire \SI{30}{\bit} instruction encoding quadrant, with a prefix of 10 for the vector processor instruction set listed in Table~\ref{tab:cust_instr}. In the remainder of this section, we highlight some key implementation details.
\begin{figure}[!t]
    \centering
    \includegraphics{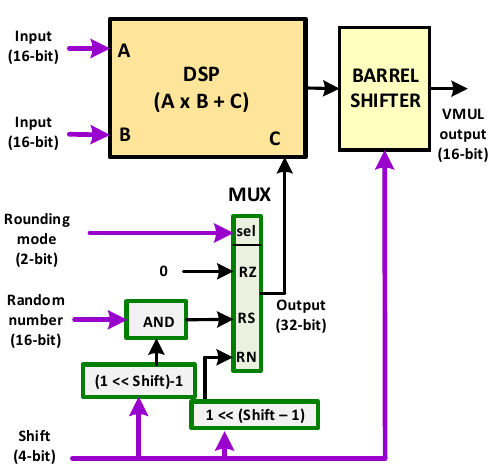}
    \caption{Execution of \textbf{VMUL} instruction in one vector lane.}
    \label{fig:mul}
\end{figure}

FeNN implements a number of specialized instructions. 
\textbf{VANDADD} performs a masked (using a mask generated from the shift encoded in funct7[3:0]) addition of a scalar and vector operand and is used when processing spikes (see Section~\ref{sec:snn_implement}).
\textbf{VRNG} produces an independent \SI{16}{\bit} random number in each lane using the Xoroshiro32++ generator~\citep{blackman_scrambled_2021} which produces relatively high-quality random numbers using only hardware-friendly addition, shift and bitwise operations.  
However, sampling from this Random Number Generator~(RNG) requires reading \SI{32}{\bit} and writing back \SI{48}{\bit} of internal state per cycle whereas, a typical RISC-V instruction can only read two operands and write one result. Therefore, if we had used standard registers to hold the RNG state, it would have required adding additional register file ports.
Instead, FeNN has two additional two-port registers to hold the RNG state. 
These are read at the start of the decode stage and a new random number is inserted into the pipeline every clock cycle.
FeNN also implements a minimal set of two-operand arithmetic instructions (\textbf{VMUL}, \textbf{VAND}, \textbf{VSUB} and \textbf{VADD}), all of which operate on signed \SI{16}{\bit} fixed-point values.
In our previous paper, we showed how saturating arithmetic can prevent catastrophic failures when neuron dynamics enter extreme activity regimes~\citep{aizaz_fenn_2025} and funct7[6] of \textbf{VADD} and \textbf{VSUB} enables this functionality.
Fixed point multiplication typically consists of a multiplication followed by a right shift, and, to support this efficiently without needing to store \SI{32}{\bit} intermediate values, \textbf{VMUL} performs a multiplication followed by a shift specified by funct7[3:0] within the one clock cycle.
As \figurename~\ref{fig:mul} shows, this is implemented with a single DSP block -- implementing a MAC operation -- and a barrel shifter per lane.
The addition input of the DSP block is used to implement three rounding modes. \textit{Round-to-zero} where zero is added, \textit{Round-to-nearest} where 0.5 (in the fixed-point format corresponding to the shift amount) is added and \textit{stochastic rounding} where the random number produced in the decode stage is added.
Additionally, FeNN implements two-operand shift instructions (\textbf{VSL} and \textbf{VSR}) as well as immediate versions  (\textbf{VSLI} and \textbf{VSRI}), which encode the shift in imm[3:0].
These immediate variants are convenient for converting between fixed point types and, to reduce rounding error when shifting right, \textbf{VSRI} additionally supports the same rounding modes as \textbf{VMUL} encoded in imm[5:4].
%
The \textbf{VTEQ}, \textbf{VTNE}, \textbf{VTLT} and \textbf{VTGE} instructions compare two vector operands and write the result to a \SI{32}{\bit} scalar register using \SI{1}{\bit} to represent the comparison result from each lane. 
These results can be used directly -- for example to store spikes -- or for masked control flow using the \textbf{VSEL} instruction, which implements a ternary operator.
%
VEC can access lane-local and vector memories.
\textbf{VLOADV} and \textbf{VSTOREV} perform \SI{64}{\byte} aligned vector memory loads and stores,
\textbf{VLOADR0} and \textbf{VLOADR1} also load from the vector memory, but write to the special RNG seed registers and
\textbf{VLOADL} and \textbf{VSTOREL} perform \SI{2}{\byte} aligned lane-local memory loads and stores.
%
Finally, FeNN provides a limited set of data movement instructions (\textbf{VEXTRACT}, \textbf{VFILL} and \textbf{VLUI}) for moving values between the scalar and vector register files.

\begin{table}%
\begin{subalgorithm}[t]{.33333\textwidth}
    \begin{algorithmic}
        \scriptsize
        \LineComment{Load weight}
        \State $\Call{VLOAD.V}{\mathbf{w}, a_w, 0}$
        \LineComment{Load target neuron inputs for next iteration}
        \State $\Call{VLOAD.V}{\mathbf{i_{next}}, a_i, 0}$
        \LineComment{Add weights to neuron inputs}
        \State $\Call{VADD.S}{\mathbf{i_{prev}}, \mathbf{i_{prev}}, \mathbf{w}}$
        \LineComment{Store target neuron inputs}
        \State $\Call{VSTORE.V}{a_i, \mathbf{i_{prev}}, 0}$
        \State
        \State
        \State
        \State
        \State
    \end{algorithmic}
    \caption{Dense}\label{alg:dense_spike_processing}
\end{subalgorithm}
\begin{subalgorithm}[t]{.33333\textwidth}
    \begin{algorithmic}
        \scriptsize
        \LineComment{Load weights \& indices for next iteration}
        \State $\Call{VLOAD.V}{\mathbf{d_{next}}, a_w, 0}$
        \LineComment{Calculate lane-local address}
        \State $\Call{VANDADD}{\mathbf{a}, \mathbf{d_{prev}}, a_i, \text{log}_2(N_\text{T})}$
        \LineComment{Load target neuron inputs}
        \State $\Call{VLOAD.L}{\mathbf{i}, \mathbf{a}, 0}$
        \LineComment{Extract weight}
        \State $\Call{VSRAI}{\mathbf{w}, \mathbf{d_{prev}}, \text{log}_2(N_\text{T})}$
        \LineComment{Add weights to neuron inputs}
        \State $\Call{VADD.S}{\mathbf{i}, \mathbf{i}, \mathbf{w}}$
        \LineComment{Store target neuron inputs}
        \State $\Call{VSTORE.L}{\mathbf{a},\mathbf{i}, 0}$
        \State
    \end{algorithmic}
    \caption{Sparse}\label{alg:sparse_spike_processing}
\end{subalgorithm}
\begin{subalgorithm}[t]{.33333\textwidth}
    \begin{algorithmic}
        \scriptsize
        \LineComment{Load weights \& delays for next iteration}
        \State $\Call{VLOAD.V}{\mathbf{d_{next}}, a_w, 0}$
        \LineComment{Calculate lane-local address}
        \State $\Call{VADD}{\mathbf{a}, \mathbf{d_{prev}}, \mathbf{t}}$
        \State $\Call{VANDADD}{\mathbf{a},\mathbf{a}, a_d, \text{log}_2(N_\text{D})}$
        \LineComment{Load target neuron inputs}
        \State $\Call{VLOAD.L}{\mathbf{i}, \mathbf{a}, 0}$
        \LineComment{Extract weight}
        \State $\Call{VSRAI}{\mathbf{w}, \mathbf{d_{prev}}, \text{log}_2(N_\text{D})}$
        \LineComment{Add weights to neuron inputs}
        \State $\Call{VADD.S}{\mathbf{i}, \mathbf{i}, \mathbf{w}}$
        \LineComment{Store target neuron inputs}
        \State $\Call{VSTORE.L}{\mathbf{a}, \mathbf{i}, 0}$
    \end{algorithmic}
    \caption{Delayed}\label{alg:delayed_spike_processing}
\end{subalgorithm}
\captionsetup{labelformat=alglabel}
\caption{Spike propagation algorithms implementing using instructions from Table~\ref{tab:cust_instr}. Variables in bold lower-case are located in vector registers, those in lower case italics are located in scalar registers and those in upper-case italics are immediate values (compile-time constants). $a_w$ holds the address of the weights (in vector memory), $a_i$ holds the  address of the target neuron inputs (either in vector or lane-local memory) and $\mathbf{t}$ holds the current simulation timestep duplicated across each vector lane. $N_\text{T}$ specifies the number of target neurons per-lane and $N_\text{D}$ the number of delay slots.}%
\label{alg:spike_processing}
\end{table}

\subsection{Implementing SNNs on FeNN}
\label{sec:snn_implement}
\subsubsection{Spiking neuron updates}
\label{sec:snn/neuron}
In discrete time, the Leaky Integrate-and-Fire~(LIF) spiking neuron model can be expressed as:
\begin{align}
    H_j^{t} &= \alpha V_j^{t-1} + I_j^t\label{eq:lif_dynamics} \\
    Z_j^t &= \begin{cases} 
        1 & \text{if } H_j^t \ge V_\text{th} \\
        0 & \text{otherwise}
    \end{cases} \label{eq:lif_threshold}\\
    V_j^{t} &= H_j^{t} - Z_j^t V_\text{th} \label{eq:lif_refractory}
\end{align}
where $V_j$ is the internal state of neuron $j$ and $I_j$ its input. 
$H_j$ is a temporary variable containing the internal state before the spiking threshold is applied, $V_\text{th}$ is the neuron's spiking threshold and $Z_j$ its binary spiking output. $\alpha$ is a decay factor which can be precomputed from the neuron's time constant $\tau$ using $\alpha=e^\frac{-1}{\tau}$.

Alternatively, LIF neurons can be updated only when spikes occur by decaying $V$ by $\alpha^{\delta t}=e^\frac{-\delta t}{\tau}$ where $\delta t$ is the time since the previous spike.
However, while this approach can \emph{potentially} save some computation, it does not generalise to more complex neuron models, whereas the equivalent of (\ref{eq:lif_dynamics}) and (\ref{eq:lif_refractory}) can be applied almost universally and also is trivially parallelisable along $j$.
Using FeNN, this parallelism is exploited by loading vectors of state variables from on-chip memory and parameters from immediates.
Then, (\ref{eq:lif_dynamics}) and (\ref{eq:lif_refractory}) can be implemented using standard arithmetic instructions.
Finally, (\ref{eq:lif_threshold}) can be implemented using FeNN's comparison instructions to write $Z$ for a whole vector of neurons to a \SI{32}{\bit} scalar register.
These vectors are then written to scalar memory.

\begin{figure}
    \centering
    \includegraphics[width=8.5cm]{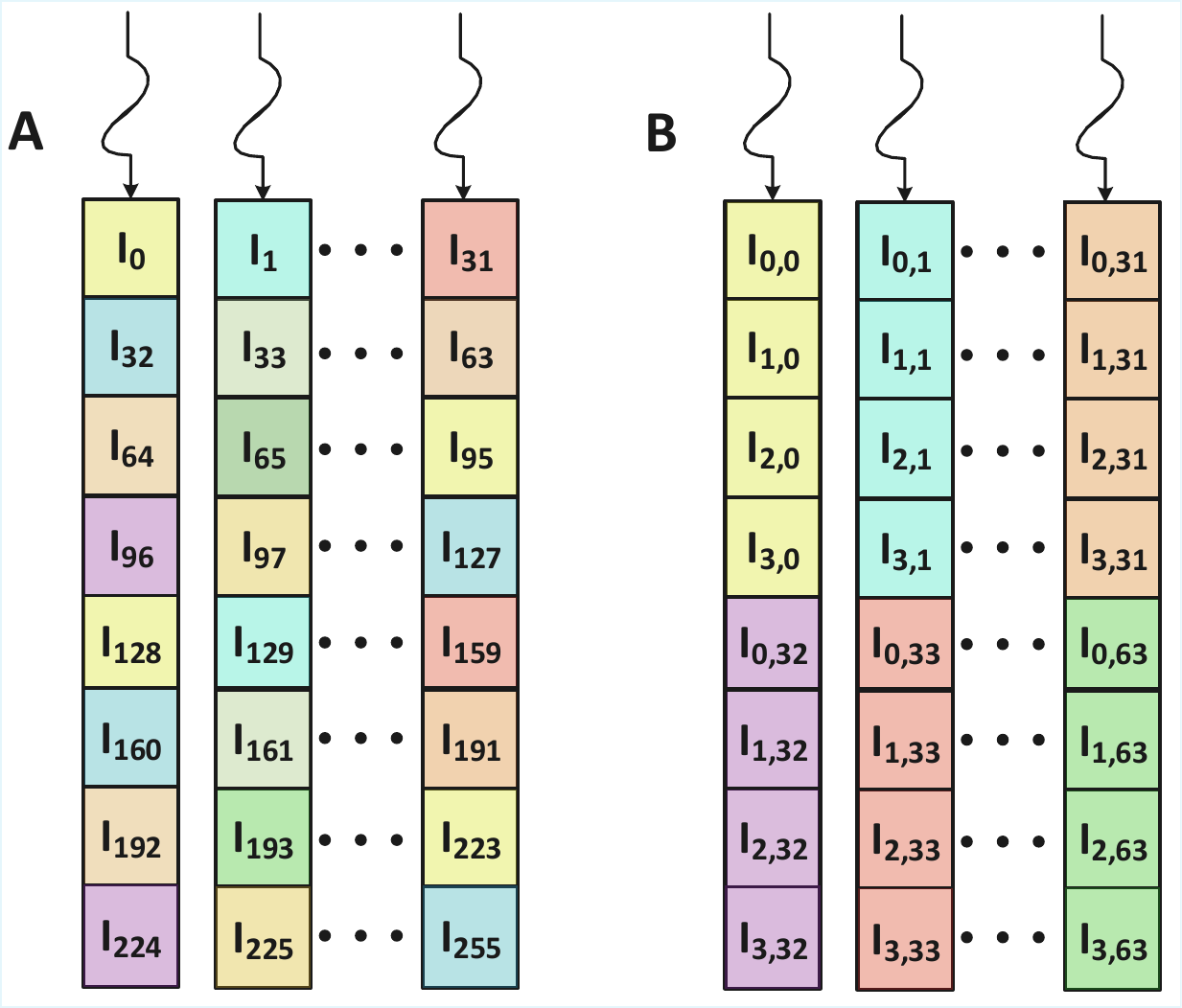}
    \caption{Lane-local memory data structures for spike propagation. Snaking lines indicate parallelism across vector lanes. Colours are used to differentiate inputs to different neurons. \textbf{(A)} Sparse connectivity with 256 target neurons. $I_i$ indicates the input to neuron $i$. \textbf{(B)} Delayed connectivity with 64 target neurons and $N_\text{D}=4$ delay slots. $I_{i,d}$ indices the input to neuron $i$ in delay slot $d$.}
    \label{fig:llm_datastructures}
\end{figure}

\subsubsection{Spike propagation}
\label{sec:snn/spike_prop}
In an SNN, the input $I_i$ to each target neuron $i$ are produced by propagating the spikes of connected neurons ($Z_j^t$) through a weight matrix ($W_{ij}$):
\begin{align}
    I_i^t &= \sum_j W_{ij} Z_j^t.
\end{align}
Because spikes are sparse, we iterate over the `ones' in the spike bitfield ($Z_j^t$) and parallelise across target neurons $i$ so contiguous `rows' of weights are accessed together -- this is vital for efficient use of both the on-chip vector memory and the DMA controller.
To efficiently iterate through the sparse bitfields containing the spikes, we first loop through the \SI{32}{\bit} words that make up the bitfield and then through the bits in each word.
The inner loop over the bits can be implemented efficiently on the scalar core by counting the leading zeros in the word (using the \textbf{CLZ} instruction from the standard `B' RISC-V extension) and then shifting them off (using the \textbf{SLL} instruction).
When using weights stored in external memory, this loop is double-buffered so that each iteration calculates the external memory address of the `row' of weights associated with a spike, launches a DMA transfer to copy them into a buffer in vector memory and -- while the data is being transferred -- propagates the \emph{previous} spike through the weights previously fetched into the second buffer.
For small models with weights stored in on-chip vector memory, double-buffering is not necessary. Vector memory addresses are calculated for each spike, and spikes are processed immediately.

As discussed in Section~\ref{sec:introduction}, processing the rows of connectivity tends to be the most costly part of an SNN simulation, and we have co-designed the FeNN instruction set alongside optimised algorithms to support dense, sparse and delayed connectivity.
Dense connectivity is handled by Algorithm~\ref{alg:dense_spike_processing}, which simply loads vectors of \SI{16}{\bit} weights and inputs to the target neuron ($I$) from vector memory, adds them together and stores back the updated inputs.
Sparse connections are encoded in \SI{16}{\bit} words, with the target neuron index in the lower bits and the weight in the upper bits.
We handle sparse connectivity using Algorithm~\ref{alg:sparse_spike_processing} (inspired by the approach described by \citet{Naylor2013}), where each lane is responsible for processing all connections to target neurons whose index modulo 32 matches the index of the lane.
Using this scheme, the $I$ values are distributed across the lane-local memories as shown in \figurename~\ref{fig:llm_datastructures}A.
Algorithm~\ref{alg:sparse_spike_processing} calculates lane-local memory addresses from these targets by simply adding the base address~($a_i$) of the data structure in lane-local memory.
Delayed connectivity is implemented in a similar manner, with delays packed into the lower bits of each \SI{16}{\bit} connection and each neuron associated with a $N_\text{D}$ element delay buffer in lane-local memory, as shown in \figurename~\ref{fig:llm_datastructures}B.
These delayed connections are processed with Algorithm~\ref{alg:delayed_spike_processing}, which builds the lane-local memory address by adding the time~($\mathbf{t}$) and the base address~($a_d$) of the data structure to the delay.

In a four-stage pipelined CPU like FeNN, data produced by a load instruction cannot be used by the ALU in the next instruction.
To avoid this causing stalls, all variants of Algorithm~\ref{alg:spike_processing} double-buffer loads so these algorithms can all run at 1 instruction per cycle.
As each iteration of these algorithms processes 32 connections, connections can be processed at a theoretical peak rate of:
\begin{align}
    T_\text{peak} &=\frac{f_\text{max} \times 32}{N}\si{\sop\per\second}\label{eq:peak_throughput}
\end{align}
where $N$ is the number of instructions in the inner loop.
This equates to a theoretical peak throughput of \SI{1.4}{\giga\sop\per\second} for dense connectivity and \SI{0.8}{\giga\sop\per\second} for delayed connectivity at \SI{175}{\mega\hertz}.
The additional complexity of Algorithm~\ref{alg:sparse_spike_processing} makes processing sparse synapses slightly slower, but this is easily compensated by the fact that fewer synapses need processing, resulting in an \emph{effective} throughput of \SI{3.72}{\giga\sop\per\second} with \SI{75}{\percent} sparsity.
Servicing these throughputs requires a maximum memory bandwidth of \SI{2.8}{\giga\byte\per\second} -- significantly lower than the \SI{4.9}{\giga\byte\per\second} achieved by our DMA controller.
This is very advantageous as it means the memory latency (measured at around 60 clock cycles) \emph{and} the time taken to transfer the next row of connectivity to vector memory can be hidden by the processing time of the current row.

\subsubsection{PyFeNN software stack}
\label{sec:methods/pyfenn}
In our previous work~\citep{aizaz_fenn_2025}, we hand-coded several example SNNs in RISC-V assembly language using the strategies outlined in the previous sections.
While this was educational, it is not a practical proposition for end-users, so we have developed a new Python-based toolchain which provides APIs for building SNNs, running them on FeNN or our behavioural simulator and interacting with simulations.
One of the key aims of FeNN is to allow different neuron models to be easily implemented by end-users and, to support this, neuron models are defined in an extended version of the C-like language used in our GeNN SNN simulation library~\citep{Knight2021}.
In GeNN, this C-like language is tokenised and then parsed using a recursive descent parser, heavily inspired by the `jlox' implementation described by \citet{nystrom_crafting_2021}.
The Abstract Syntax Tree~(AST) this parser produces is then type-checked and `pretty-printed' into CUDA or C++ to run on standard hardware.
In PyFeNN, we have extended the type checker to support fixed-point types and currently use a simple single-pass compiler of the type described by \citet{wirth_compiler_1996} to directly generate vectorised FeNN from the AST.

Using our PyFeNN toolchain, the time-driven update of a simple LIF neuron defined in (\ref{eq:lif_dynamics}), and (\ref{eq:lif_threshold}) can be implemented using a \lstinline[language=Python]{NeuronUpdateProcess} which is implemented on FeNN using the strategy described in Section~\ref{sec:snn/neuron}:

\begin{lstlisting}[language=Python]
class LIF:
    def __init__(self, shape, tau_m, v_thresh):
        self.shape = shape
        self.v = Variable(self.shape, "s7_8_sat_t")
        self.i = Variable(self.shape, "s7_8_sat_t")
        self.out_spikes = EventContainer(self.shape)

        self.process = NeuronUpdateProcess(
            """
            V = (Alpha * V) + I; 
            I = 0;
            if(V >= VThresh) {
               Spike();
               V -= VThresh;
            }
            """,
            {"Alpha": Parameter(np.exp(-1.0 / tau_m), 
                                "s0_15_sat_t"),
             "VThresh": Parameter(v_thresh, 
                                  "s7_8_sat_t")},
            {"V": self.v, "I": self.i}, 
            {"Spike": self.out_spikes})
\end{lstlisting}
where \lstinline[language=Python]{Variable} objects encapsulate arrays of \SI{16}{\bit} values and are either allocated in vector or lane-local memory (depending on whether they are used to hold the output of event propagation through sparse connectivity).
\lstinline[language=Python]{EventContainer} objects encapsulate the bitfield arrays used to store events and are always stored in scalar memory.

\lstinline[language=Python]{NeuronUpdateProcess} is just one example of `processes' which represent tasks to be offloaded to FeNN. 
Others include \lstinline[language=Python]{EventPropagationProcess} which encapsulates the various means to propagate spikes between neurons described in Section~\ref{sec:snn/spike_prop} and a number of utility processes, such as \lstinline[language=Python]{MemsetProcess} and \lstinline[language=Python]{BroadcastProcess} used for initialising variables in on-chip memory.
Similarly to the \lstinline[language=Python]{LIF} example above, PyFeNN encapsulates these objects in Python classes, so a simple SNN with one hidden layer can be defined as:
\begin{lstlisting}[language=Python]
# Input spikes
input_spikes = EventContainer(input_shape, 
                              num_timesteps)
# Neurons
hidden = LIF(hidden_shape, 20.0, 1.0)
output = LI(output_shape, 20.0)
# Synapses
input_hidden = Linear(input_spikes, hidden.i)
hidden_output = Linear(hidden.out_spikes, output.i,
                       "s9_6_sat_t")
\end{lstlisting}
PyFeNN also supports the Neuromorphic Intermediate Representation~\citep{pedersen_neuromorphic_2024}, allowing models trained using a wide range of software libraries to be easily deployed on FeNN.
\setlength\rotFPtop{200pt}
\begin{sidewaystable}
\scriptsize
\begin{tabular}{lcccccc}
\toprule
 &\citet{dvs} &\citet{fpga_nhap} & \citet{essa} & \citet{spiker+}$^\dagger$ & \citet{li_fully-parallel_2024} & \textbf{FeNN-DMA}\\
\midrule
 Device  &ZCU104 &  Kintex-7& Kintex UltraScale  &XCZU3EG & ZCU102 & Kria KV260 \\
 Frequency (\si{\mega\hertz}) &250 &200 &140 & 100& 30 &175\\

    
Max. neurons &\num{4096}  &\num{16000}  &\num{2048} &\num{1900}&\num{2500}&\num{16000} \\

Max. synapses (thousands) &\num{1000} &\num{16800} & \num{512} &--& \num{4.9}&\num{256000}\\

 Datasets & N-MNIST & MNIST& MNIST & MNIST & MNIST & N-MNIST \\
  & DVS gesture & & &  AudioMNIST && SHD \\
  &  & &  &  SHD && \\
  
  Off-chip memory  &No   & Yes & No  &  No & No &Yes\\
  Delays  &No   & Yes & No  &  No & No & No\\
  Recurrent &No & No & No  & Yes  & No  & Yes \\
  Weight compression & Structured & No & Unstructured& No & Structured & Unstructured\\
Neuron model  &LIF & LIF, Izhikevich &LIF & IF, LIF, CUBA-LIF &IF & Programmable\\

Dynamic power (W) &0.808$^\ddagger$ & 0.535 & 1.3$^\ddagger$ & 1.2 & 1.18& 0.53 \\

Dense throughput (\si{\giga\sop\per\second}) & 3.49 & 0.019 &68.2 &-- & 47.3 & 0.92\\
LUTs & \num{81299} & \num{46371} & \num{585978}  &  \num{62989} & \num{174362} & \num{54984}\\
FFs & \num{47768} & \num{30417} & \num{232686}  & --  & \num{95000} & \num{47759}\\
BRAMs & \num{258.5} & \num{150} & \num{432}  & \num{215}  & -- & \num{66}\\
\bottomrule
\end{tabular}
\caption{Comparison of single-core FeNN-DMA SoC to state-of-the-art FPGA-based SNN accelerators.\\
$^\dagger$ For a fairer comparison with larger systems, this columns refers to the largest presented Spiker+ architecture.\\
 $^\ddagger$ These power estimates are produced using a more accurate methodology based on switching activity files.}
 \label{tab:hardware_comparison}
\end{sidewaystable}
\section{Results}
\label{sec:results}
\subsection{Implementation}
FeNN-DMA was developed in SystemVerilog and we synthesised and implemented one and two-core FeNN-DMA SoCs using the AMD Vivado Design Suite 2023.2. 
By carefully optimising critical paths and employing a range of strategies such as Flow\_PerfOptimized\_high for synthesis, Performance\_ExtraTimingOpt for implementation, and floorplanning, we achieved an operating frequency of \SI{175}{\mega\hertz} for our single-core design which is very competitive with other softcore vector processors, even those with much narrower vector units~\citep{kuo_integration_2023}. 
We also synthesised a dual-core FeNN SoC which required approximately double the LUT, FF and BRAM resources of the single-core SoC and achieved a reduced operating frequency of \SI{143}{\mega\hertz} due to routing congestion.

In Table~\ref{tab:hardware_comparison}, we compare FeNN to the state-of-the-art systems described in Section~\ref{sec:introduction}. It is clear that throughput is strongly correlated with the amount of available parallelism and, hence, the resource usage.
Thus, while the very large systems~\citep{essa} and those with dedicated circuits for each neuron~\citep{li_fully-parallel_2024} achieve much higher throughputs than FeNN, they are also many times bigger while supporting far fewer neurons and synapses.
Similarly, the largest accelerator generated by the Spiker+ framework~\citep{spiker+} uses more resources than FeNN yet can only simulate a fraction of the number of neurons.

The fully event driven system developed by \citet{dvs} provides the most relevant comparison point as it uses time multiplexing and, while it has half the parallelism of FeNN (16 PEs), each PE runs at a higher clock-speed (\SI{250}{\mega\hertz}) and implements a fixed-function pipeline capable of processing one synaptic operation each clock cycle.
FeNN is unable to operate at such high clock speeds as the forwarding logic between the writeback and decode pipeline (see Section~\ref{sec:methods/cv32e40x} and Section~\ref{sec:methods/vec}), which is unnecessary in a fixed-function system, lengthens the critical paths and reduces the maximum clock frequency.
Furthermore, as discussed in Section~\ref{sec:snn/spike_prop}, on FeNN, each synaptic operation takes a minimum of 4 clock cycles. These factors combine to give the system developed by \citet{dvs} a $3\times$ throughput advantage.

The additional IPs required to access external memory add an overhead of around \SI{32}{\percent} in terms of LUTs and FlipFlops to the FeNN-DMA SoC.
However, the resource usage of the single-core FeNN-DMA SoC is still  remarkably similar to the fixed-function system although FeNN's support for external memory enables it to simulate many more neurons and synapses than \emph{any} other system.
This demonstrates the advantages of wide vector architectures on FPGAs, as the cost of programmability, i.e. the RISC-V host processor, is amortised across many vector lanes.
Furthermore, this wide architecture allows each FeNN core to use 16 high-density URAM memories to implement its vector memories rather than the BRAMs which are the limiting resource in many other state-of-the-art designs.
Accurately measuring the power usage of FeNN-DMA is difficult. 
The Kria KV260 development board only exposes a single INA3221 power monitor connected to the power rail driving the entire SoC, meaning there is no way to separate the power usage of FeNN-DMA running on the PL from the quad-core ARM Cortex-A53 CPU, Mali GPU etc in the PS.
Additionally, because FeNN is designed to be controlled from Ubuntu running on the PS, it is impossible to perform meaningful post place-and-route simulations of the whole system and thus generate the switching activity files required to produce a more accurate power estimation.
Instead, the best we can do is perform a `vectorless' power estimation using Vivado and remove the estimated power use of the PS (ARM Cortex-A53 CPU, Mali GPU etc).
Measured in this way, the single-core SoC requires a dynamic power of \SI{0.53}{\watt} and the dual-core around \SI{0.70}{\watt}.
These values are not dissimilar to fixed-function systems with similar resource usage.

\begin{figure}
    \centering
    \includegraphics{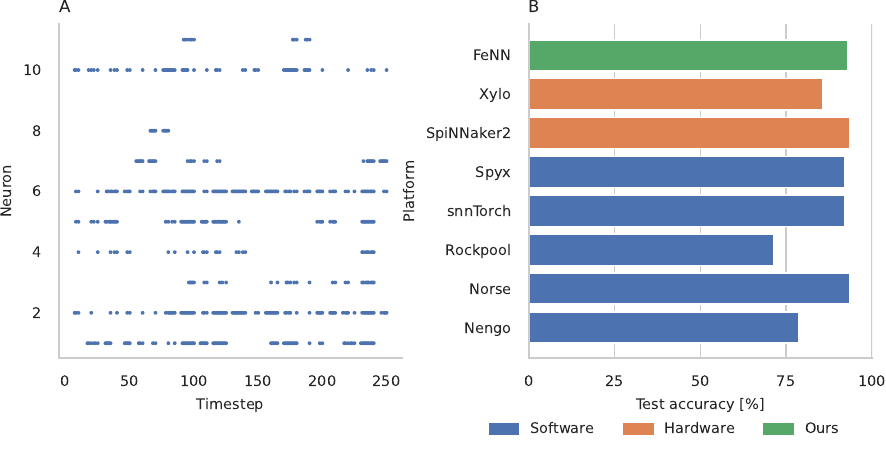}
    \caption{Spiking Recurrent Neural Network for tactile classification of Braille digits~\citep{muller-cleve_braille_2022}. \textbf{(A)} Example raster plot showing one sample from the dataset. \textbf{(B)} Classification accuracy compared to other NIR backends tested by \citet{pedersen_neuromorphic_2024}.}
    \label{fig:nir}
\end{figure}
\subsection{Spiking Neural Network classifiers}
Adding trainable synaptic delays to SNNs has recently been shown to improve their performance in spatio-temporal tasks such as keyword recognition~\citep{hammouamri_learning_2023,meszaros_efficient_2025}.
Here, we deploy one of the networks trained by \citet{meszaros_complete_2025} on the Spiking Heidelberg Digits~(SHD) dataset~\citep{Cramer2020} onto FeNN.
This model has a single, recurrently connected hidden layer of 256 LIF neurons, and all recurrent connections and input to the hidden connections have individual delays of between 0 and 62 timesteps.
Running on FeNN, this model obtains an accuracy of \SI{90.32(0.1)}{\percent} on the SHD test set, and each digit takes, on average, \SI{10.1}{\milli\second} to classify (\SI{8.6}{\micro\second} per timestep).
The trainable synaptic delays in this model result in a large accuracy improvement compared to Spiker+~\citep{spiker+} (the only other FPGA-based accelerator we are aware of that was evaluated on SHD), which only achieved \SI{72.99}{\percent}, although at a lower latency of \SI{5.4}{\micro\second} per timestep.
Compared to the latencies reported by \citet{meszaros_complete_2025} for the same model running on Loihi 2 and a Jetson Orin Nano, FeNN operates at half the latency of the Jetson but around $7\times$ the latency of Loihi 2 (this is unsurprising as Loihi 2 is a large commercial ASIC developed on an advanced Intel 4 process node).
While we have not yet developed an interconnect network for multi-core FeNN-DMA SoCs -- which would allow networks to be distributed between cores and thus reduce latency -- we can use our dual-core SoC for \emph{data parallel} inference.
This reduces the total classification time of the SHD test set by \SI{40}{\percent}.

\citet{dvs} trained models with structured sparsity on the Neuromorphic-MNIST dataset~\citep{orchard_converting_2015} and performed inference using their event-based FPGA accelerator.
They reported accuracies ranging from \SI{98.11}{\percent} for a dense network down to \SI{96.88}{\percent} for a version using connectivity with \SI{75}{\percent} structured sparsity.
To demonstrate the advantages of the \emph{unstructured sparsity} and recurrent connectivity supported by FeNN, we trained an extremely sparse model consisting of a single hidden layer of \num{512} LIF neurons, recurrently connected with \SI{99}{\percent} sparsity and connected to the $34\times34\times2$ N-MNIST input with \SI{95}{\percent} sparsity.
We trained this model using the approach described in our recent work~\citep{knight_flexible_2025} and, when deployed onto FeNN, this model achieves an accuracy of \SI{98.46(0.02)}{\percent} on the N-MNIST test set -- higher than the dense model presented by \citet{dvs} and with around $4\times$ fewer parameters than their sparsest model. 

Finally, we evaluate the Neuromorphic Intermediate Representation~(NIR) importer described in Section~\ref{sec:methods/pyfenn} using the simplified Braille tactile classification task described by \citet{pedersen_neuromorphic_2024}.
Quantized and running on FeNN, the classifier trained by \citeauthor{pedersen_neuromorphic_2024} obtains an accuracy of \SI{92.86}{\percent} on the test set.
As Figure~\ref{fig:nir}B shows, this is very similar to the accuracy obtained by the best software backends.

\begin{figure}
    \centering
    \includegraphics{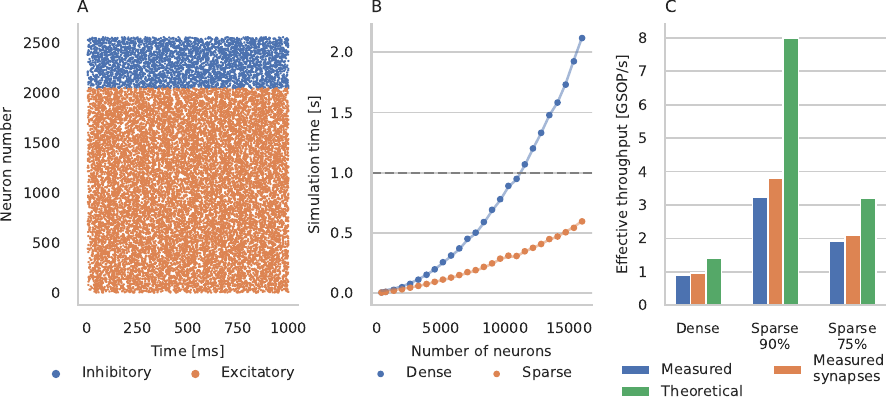}
    \caption{\textbf{(A)} Example raster plot showing activity of neurons in a balanced random network with 2048 excitatory and 512 inhibitory neurons. \textbf{(B)} Simulation time of a balanced random network with \SI{90}{\percent} sparsity running on a single FeNN core. Points represent measured simulation times, and lines in corresponding colours show the time predicted by our performance model based on the number of neurons and number of SOPs. The horizontal dashed line represents real-time performance. \textbf{(C)} Effective throughput of balanced random network with \num{16000} neurons. ``Measured'' is based on total simulation time, ``Measured synapses'' is calculated using performance counters around the event propagation process group and ``Theoretical'' is calculated as described in Section~\ref{sec:snn_implement}.}    
    \label{fig:va_benchmark}
\end{figure}

\subsection{Performance scaling}
While the classifier results presented in the previous section are indicative of FeNN's performance in real applications, it is difficult and costly to train such models at a wide enough range of scales to fully explore the performance of an SNN accelerator.
Instead, we employed a standard model from the computational neuroscience literature known as a \emph{Balanced Random Network} consisting of two recurrently connected populations of spiking neurons.
Specifically, we used the widely used benchmark model of \citet{Vogels2005}.
The parameter values for a fixed spiking rate in the network (see \figurename~\ref{fig:va_benchmark}A for example model activity) can be calculated empirically for models of any size and sparsity.
\figurename~\ref{fig:va_benchmark}B shows how the time taken to simulate such a network with \SI{90}{\percent} sparse connectivity for \num{1000} \SI{1}{\milli\second} timesteps on a single-core FeNN-DMA SoC scales with the number of neurons.
Using dense connectivity, up to \num{10000} neurons can be simulated in real-time on each core and, with sparse connectivity, we can simulate up to \num{16000} neurons at faster than real-time before running out of lane-local memory.

As well as recording the total simulation time, we also record all the emitted spikes and the number of clock cycles spent propagating events and updating neurons using RISC-V performance counters.
\figurename~\ref{fig:va_benchmark}C shows these measurements for the largest network size (\num{16000} neurons) as well as for a network with \SI{75}{\percent} sparse connectivity for better comparison with \citet{dvs}.
Clearly, at this scale, the total simulation time (`Measured') is dominated by the time spent propagating events (`Measured synapses'), suggesting that there would be minimal advantage to fully event-based neuron updates.
The measured throughput of dense synapse processing is around \SI{70}{\percent} of the theoretical peak, calculated using (\ref{eq:peak_throughput}), suggesting that, unsurprisingly, iterating through spikes and initiating DMA transfers in software is less efficient than a pure hardware system (\citet{dvs} achieve around \SI{91}{\percent} of their theoretical peak performance).
However, the larger gap between the theoretical and measured throughput for sparse connectivity is due to the randomly sampled connections not being evenly distributed between the lanes, meaning that they cannot be perfectly compressed by the data structure described in Section~\ref{sec:snn/spike_prop}.
In fact, almost half the entries in each row of connectivity are zeroes, inserted for padding.
Nonetheless, as reported by \citet{dvs}, even though propagating spikes through sparse connectivity is less efficient (see Section~\ref{sec:snn_implement}), it does significantly increase \emph{effective} spike processing throughput ($3.2\times$ with \SI{90}{\percent} sparsity and $2.1\times$ with \SI{75}{\percent} sparsity.

Both \citet{chen_gaban_2022} and \citet{Naylor2013} benchmarked their programmable SNN accelerators on similar balanced random networks.
\citet{chen_gaban_2022} simulated a network of \num{10000} LIF neurons firing at a higher rate, with \SI{99}{\percent} sparse connectivity.
Simulating this model for one biological second on FeNN takes \SI{0.36}{\second} compared to \SI{3.2}{\second} on a single GABAN core -- probably due to a combination of FeNN's wider SIMD unit and the GABAN simulation perhaps including STDP.
\citet{Naylor2013} simulated a larger (\num{64000} neurons) but less densely connected network of slightly more complex Izhikevich~\citep{Izhikevich2004} neurons firing at a higher rate.
FeNN's lane-local memories are not large enough to simulate sparse models with this many neurons on a single core.
Instead, we estimated performance on this model by fitting 1\textsuperscript{st} order polynomials to the performance data we previously collected, specifically the number of clock cycles spent updating $N_\text{neuron}$ neurons and propagating $N_\text{sop}$ sparse synaptic events.
Using these two polynomials, we can model the simulation time $T$:
\begin{align}
    T &= \frac{(a_\text{neuron} + b_\text{neuron} N_\text{neuron}) + (a_\text{synapse} + b_\text{synapse} N_\text{sop})}{f_\text{max}}
\end{align}
where $a_\text{neuron}$ and $b_\text{neuron}$ are coefficients fitted to the neuron update data; and $a_\text{synapse}$ and $b_\text{synapse}$ are coefficients fitted to the event propagation data.
\figurename~\ref{fig:va_benchmark}B illustrates how well this model fits our data and, after multiplying $b_\text{neuron}$ by \num{2.6} -- reflecting the increased cost of updating Izhikevich neurons estimated by  \citet{Izhikevich2004} -- this estimate suggests a single FeNN core could simulate this model for one biological second in \SI{1.83}{\second} compared to \SI{3.9}{\second} for a single BlueVec core -- largely reflecting FeNN's wider SIMD unit.


\section{Discussion}
\label{sec:conclusion}
Here we have presented our FeNN-DMA architecture.
FeNN-DMA is a RISC-V-based SNN accelerator with an instruction set customised to the needs of SNN simulation and designed for FPGA deployment.
We have demonstrated FeNN-DMA on several challenging spatio-temporal benchmarks and shown that, due to its flexible support for cutting-edge SNN architectures with recurrent connectivity and delays, we can achieve significantly higher accuracy than other FPGA-based accelerators.
Furthermore, due to our customised DMA controller, FeNN can stream weights from off-chip memory with minimal performance overhead.
This not only allows us to simulate very large models, but also reduces the pressure to aggressively reduce weight precision.
Therefore, models can be trained using standard approaches and the weights quantised to \SIrange{8}{16}{\bit} fixed-point using Post Training Quantization.

FeNN-DMA's power and resource requirements are not dissimilar to those of equivalent fixed-function FPGA-based accelerators but, there is a throughput gap of $3-4\times$ compared to these fixed-function systems.
This is due to our software spike propagation loop (see Algorithm~\ref{alg:spike_processing}), which currently requires between 4 and 7 clock cycles to process a vector of 32 weights (depending on delays and sparsity).
However, if weights are streamed from external memory using the DMA controller, optimising Algorithm~\ref{alg:spike_processing} would not actually help throughput as the current ratio of external memory and processing throughput are well balanced to be entirely hide external memory latency (see Section~\ref{sec:snn/spike_prop}).
Nonetheless, if we were to produce customised versions of FeNN using only on-chip memory, there is potential to implement Algorithm~\ref{alg:spike_processing} directly in fixed-function pipelines within FeNN's ALU and load-store unit, which would enable them to process a new vector of weights every clock cycle.
Furthermore, because each instruction in Algorithm~\ref{alg:spike_processing} is handled by different units in the ALU and load-store unit, these pipelines could simply connect together existing hardware units, so resource overheads could be minimised.

We have already demonstrated that a dual-core FeNN-DMA design can be instantiated on the K26 SoM but, the next vital step will be to connect the cores together so spikes can be transmitted between them and even larger models can thus be simulated.
Once this interconnect is in place, we will connect a Metavision Starter Kit KV260 event-based camera, turning FeNN-DMA into a complete spiking vision system

One vital area we have not yet investigated is applying our architecture to \emph{training} SNNs.
Many past accelerator designs have implemented Spike-Timing-Dependent Plasticity~(STDP)~\citep{euler_runge,morphbungee} for training on-chip, but this does not scale to more complex networks and does not reach the accuracy of models trained using gradient-based methods.
Instead, we plan to implement Eventprop~\citep{wunderlich_event-based_2021} -- an event-based version of Backpropagation Through Time which has the same computational properties as SNN inference and thus will be well-suited to acceleration on FeNN.
Because, like all gradient-based methods, EventProp require large training datasets, we will further scale up the FeNN architecture to larger FPGAs like Alveo U55C accelerators to allow batch-parallel Eventprop training.

\section*{Code availability}
All simulations were performed using the PyFeNN 0.02 release available from \url{https://github.com/neworderofjamie/riscv_ise/}.
This release comes with bitstream suitable for use on a Kria KV260 development kit.

\section*{Acknowledgements}
 This work was funded by EPSRC grants EP/V052241/1 and EP/S030964/1; and the EU’s Horizon 2020 research and innovation programme under Grant Agreement 945539. Hardware and licenses were provided by the Xilinx University Program.
\printbibliography

\end{document}